\documentclass{article}
\usepackage{ijcai15}
\usepackage{amsmath}
\usepackage{longtable}
\usepackage{graphicx}
\usepackage{epstopdf}
\usepackage{times}
\usepackage{stfloats}
\usepackage{caption}
\usepackage{multirow}
\newtheorem{theorem}{Theorem}[section]

\newtheorem{proposition}[theorem]{Proposition}

\newenvironment{proof}[1][Proof]{\begin{trivlist}
\newcommand{\qed}{\nobreak \ifvmode \relax \else
      \ifdim\lastskip<1.5em \hskip-\lastskip
      \hskip1.5em plus0em minus0.5em \fi \nobreak
      \vrule height0.75em width0.5em depth0.25em\fi}
\item[\hskip \labelsep {\bfseries #1}]}{\end{trivlist}}

\begin{document}

\title{Learning Better Word Embedding by Asymmetric Low-Rank Projection of Knowledge Graph}
\author{\normalsize Fei Tian\\
\normalsize University of Science and Technology of China\\
\normalsize tianfei@mail.ustc.edu.cn
\And
\normalsize Bin Gao\\
\normalsize Microsoft Research\\
\normalsize bingao@microsoft.com
\AND
\normalsize Enhong Chen\\
\normalsize University of Science and Technology of China\\
\normalsize cheneh@ustc.edu.cn
\And
\normalsize Tie-Yan Liu\\
\normalsize Microsoft Research\\
\normalsize tyliu@microsoft.com
}

\maketitle \vspace{-3em}

\begin{abstract}
  Word embedding, which refers to low-dimensional dense vector representations of natural words, has demonstrated its power in many natural language processing tasks. However, it may suffer from the inaccurate and incomplete information contained in the free text corpus as training data. To tackle this challenge, there have been quite a few works that leverage knowledge graphs as an additional information source to improve the quality of word embedding. Although these works have achieved certain success, they have neglected some important facts about knowledge graphs: (i) many relationships in knowledge graphs are \emph{many-to-one}, \emph{one-to-many} or even \emph{many-to-many}, rather than simply \emph{one-to-one}; (ii) most head entities and tail entities in knowledge graphs come from very different semantic spaces. To address these issues, in this paper, we propose a new algorithm named ProjectNet. ProjecNet models the relationships between head and tail entities after transforming them with different low-rank projection matrices. The low-rank projection can allow non \emph{one-to-one} relationships between entities, while different projection matrices for head and tail entities allow them to originate in different semantic spaces. The experimental results demonstrate that ProjectNet yields more accurate word embedding than previous works, thus leads to clear improvements in various natural language processing tasks.
\end{abstract}

\section{Introduction}
\label{introduction}
In recent years, the research on word embedding (or distributed word representations) has made promising progresses in many natural language processing tasks \cite{bengio2003neural,Scratch2008,EricHuang2012,W2VFirst,NIPS2013W2V,Glove}. Different from traditional one-hot discrete representations of words, word embedding vectors are dense, continuous, and low-dimensional. They are usually trained with neural networks on a large-scale free text corpus, such as Wikipedia, news articles, and web pages, in an unsupervised manner.

While word embedding has demonstrated its power in many circumstances, it is gradually recognized that conventional word embedding techniques may suffer from the incomplete and inaccurate information contained in the free text data. On one hand, due to the restrictive topics and coverage of a text corpus, some words might not have sufficient contexts and therefore might not have reliable word embeddings. On the other hand, even if a word has sufficient contextual data, the free texts might be inaccurate, thus might not provide a semantically precise view of the word. As a result, the learned word embedding might be unable to carry on the desirable semantic information. To tackle this problem, recently some researchers have proposed to leverage knowledge graphs, such as WordNet \cite{WordNet} and Freebase \cite{Freebase08}, as additional data sources to improve word embedding \cite{JiangBinECML,KnowTextJoint,RCNet}.

A knowledge graph contains a set of nodes representing entities and a set of edges corresponding to the relationships between entities. In other words, a knowledge graph can be regarded as a set of triples $(h,r,t)$, where head entity $h$ and tail entity $t$ share the relationship $r$. In \cite{RCNet,KnowTextJoint}, in addition to the original likelihood loss on the free texts, an extra loss function is imposed to capture the relationships in the knowledge graph. Specifically, the additional loss takes the form $L_{\mathcal{K}}=\sum_{(h,r,t)}||\mathbf{h} + \mathbf{r}-\mathbf{t}||_2^2$, where $\mathbf{h}, \mathbf{t} $ are the embedding vectors of the words (entities) $h$ and $t$ respectively, and $\mathbf{r} $ is the embedding of the relationship $r$.\footnote{Please note that throughout the paper we will use bold characters to represent the embedding vectors for corresponding items.} Then the embeddings are learned by minimizing the overall loss on both free text and knowledge graph.


While the above approaches have shown certain success, we would like to point out their limitations.

First, the loss function $L_{\mathcal{K}}$ in these works cannot capture complex relationships between entities. In particular, it will encounter problem when the relationships are \emph{one-to-many}, \emph{many-to-one}, or \emph{many-to-many}. For example, $r=$``cause of death'' is a \emph{many-to-one} relationship, since many different head entities $h_i$ (e.g., $h_1$ = ``Abraham Lincoln'' and $h_2$ = ``John F Kennedy'') correspond to the same tail entity (e.g., $t = $``assassination by firearm''). In this case, the minimization of $L_{\mathcal{K}}$ will enforce the embedding vectors of all head entities (e.g., $\mathbf{h_1}$, $\mathbf{h_2}$) to approach each other, which is clearly unreasonable.

Actually such kind of complex relationships are very common in knowledge graphs. Take a widely used benchmark dataset \emph{FB13} \cite{Socher2013NTN}, which is a subset of Freebase, as an instance. For every relationship in \emph{FB13}, we calculate the average number of head entities corresponding to one tail entity and the average number of tail entities corresponding to one head entity. Then we obtain the means and standard deviations of such values under different relationships. The overall statistical information is listed in Table \ref{tbl_statis_result}, from which we can see that relationships in \emph{FB13} are highly non \emph{one-to-one}, especially for the mapping from tail entity to head entity, as shown by the large mean value of \emph{\#Head per Tail}. In addition, the standard deviation for \emph{\#Head per Tail} is fairly large, indicating that the degrees of non \emph{one-to-one} mappings from tail entity to head entity vary drastically across different relationships. This clearly shows that the issue is very serious and we should tackle it in order to learn a reasonable word embedding.

\begin{table}[htpb]
\setlength{\abovecaptionskip}{+0.1cm}
\setlength{\belowcaptionskip}{-0.2cm}
\footnotesize
\centering

\begin{tabular}{|c|c|c|c|}
 \hline
 \multicolumn{2}{|c|}{\textbf{\#Tail per Head}} &
 \multicolumn{2}{c|}{\textbf{\#Head per Tail}} \\
 \hline
 Mean & Std.Deviation & Mean & Std.Deviation \\
 \hline
 1.26 & 0.23 & 2614.17 & 9229.75 \\
 \hline
 \end{tabular}
\caption{Number of head entities per tail entity and number of tail entities per head entity in \emph{FB13}.}
\label{tbl_statis_result}
\end{table}

Second, the loss function $L_{\mathcal{K}}$ adopts simple arithmetic operations on the embedding vectors of the head and tail entities, implying that both entities are located in the same space. However, the fact is that head entities are usually more concrete and tail entities are more abstract, making it unreasonable to simply regard them as in a homogeneous space. Still use the above example, for the relationship $r = $``cause of death'', all the head entities are real human names whereas all the tail entities are abstract reasons of death. What's more, according to Table \ref{tbl_statis_result}, head and tail entities are not symmetric from the statistics perspective: the number of tail entities per head entity is much smaller than that of head entities per tail entity, further indicating the heterogeneity nature of head and tail entities and suggesting that we should treat them separately in the mathematical modeling.

In the literature, there are some research works that try to resolve one of the aforementioned issues, however, as far as we know, none of the works successfully addressed both issues. For example, in \cite{TransH}, it is proposed to project the embedding vectors of both entities onto a relation-dependent hyperplane before computing the loss function $L_{\mathcal{K}}$. However, the heterogeneity between head and tail entities is not considered. Furthermore, the projection matrix used in \cite{TransH} has a fixed rank for all types of relationships, which could not express various degrees of non \emph{one-to-one} mappings. In \cite{BordesAAAI}, different transformations are adopted to head and tail entities respectively, however, no consideration is taken to address the issue of non \emph{one-to-one} mappings.

To address the limitations of existing works, in this paper, we propose a new algorithm called ProjectNet, which adopts different and carefully designed projections to the head and tail entities respectively when defining the loss function $L_{\mathcal{K}}$. First, we show that the necessary condition to resolve the issue of non \emph{one-to-one} mapping is to ensure the projection matrix to be low-rank. In such a way, we can guarantee the translation distance between the entities to be small after projection without forcing their embedding vectors to be the same. Actually, it can be proven that the TransH model in \cite{TransH} is our special case in the sense that it also adopts a projection matrix of low (and fixed) rank. Our model is more general since we can explicitly control the rank of the projection matrix, so as to adapt to knowledge graphs with different degrees of non \emph{one-to-one} mappings. Second, by using different projection matrices for head and tail entities respectively, we can avoid the homogeneity assumption on the semantic space and therefore build a more flexible and accurate model. For example, for the knowledge graph \emph{FB13}, we should adopt a low-rank projection matrix for head entities since the number of head entities is very large for each tail entity; however, it is safe to use a relatively full rank projection matrix for tail entities since the number of tail entities is rather small for each head entity.

We have tested the performance of our proposed algorithm on several benchmark datasets, and the experimental results show that our proposal can significantly outperform the baseline methods. This indicates the benefit of carefully modeling entities and relationships when incorporating knowledge graphs into the learning process of word embedding.

The rest of the paper is organized as following. In Section \ref{related}, we summarize related works in leveraging knowledge graph to help word embedding. Then in Section \ref{model}, the detailed model is introduced and its difference with related methods is illustrated. After that, the experimental settings and results are reported in Section \ref{exp}. The paper is finally concluded in Section \ref{Conclusion}.

\section{Related Work}
\label{related}
Word embeddings, (a.k.a. distributed word representations) are usually trained with neural networks by maximizing the likelihood of a text corpus. Based on several pioneering efforts \cite{bengio2003neural,Scratch2008,Turian2010word}, the research works in this field have grown rapidly in recent years \cite{EricHuang2012,W2VFirst,NIPS2013W2V,Glove,MorphmeICML14}. Among them, \emph{word2vec} \cite{W2VFirst,NIPS2013W2V} draws quite a lot of attention from the community due to its simplicity and effectiveness.  An interesting result given by \emph{word2vec} is that the word embedding vectors it  produces can reflect human knowledge via some simple arithmetic operations, e.g., $v(Japan)-v(Tokyo)\approx v(France)-v(Pairs)$.

However, as aforementioned, word embedding models like \emph{word2vec} usually suffer from the incompleteness and inaccuracy of the free-text training corpus. To address this challenge, there are some attempts that leverage additional structured or unstructured human knowledge to enhance word embeddings. Here are some examples. In \cite{Luong13_morpheme,CuiGBQL14}, the authors adopted morphological knowledge to aid the learning of rare words and new words. In \cite{YuACL14}, the authors used semantic relational knowledge between words as a constraint in learning word embedding vectors. In \cite{RCNet}, the authors leveraged knowledge graphs, the most widely used structured knowledge, to help improve word representations. In particular, the authors did not only minimize the loss on the text corpus, but also minimized the loss on the knowledge graph by sharing embedding vectors between words and entities. In \cite{KnowTextJoint}, the authors proposed a very similar method to \cite{RCNet}, but with a different objective of improving knowledge graph understanding with the help of text corpus. Actually, both the models in \cite{RCNet} and \cite{KnowTextJoint} are inspired by the TransE model \cite{TransE}, which is a state-of-the-art work in the literature of computing distributed representations for knowledge graphs \cite{BordesAAAI,Latent4Know2012,Socher2013NTN}. In TransE,  the relational operation between entities $h$, $t$ with relationship $r$ is assumed to be a simple linear translation, i.e., $\min||\mathbf{h}+\mathbf{r}-\mathbf{t}||_2^2$. However, as pointed out in the introduction, such a simple formulation cannot handle the non \emph{one-to-one} mappings between entities. To tackle the problem, in \cite{TransH}, the authors proposed a simple projection method named TransH. We will review the detailed mathematical forms of these models in Section \ref{relation} and discuss their relationship with our proposal.

\section{The ProjectNet Algorithm}
\label{model}
In this section, we introduce our proposed ProjectNet model in details. In general, following \cite{RCNet,KnowTextJoint}, given a training text corpus $\mathcal{D}$ and a set $\mathcal{K}$ of triples in the form (\emph{head entity, relation, tail entity}) extracted from a knowledge graph, our model jointly minimize a linear combination of the loss items on both text and knowledge:
\begin{equation}
\label{eqn_total_loss}
L = \alpha L_{\mathcal{D}} + (1 - \alpha) L_{\mathcal{K}},
\end{equation}
where $\alpha\in[0,1]$ is used to trade off the two loss terms. $L_{\mathcal{D}}$ and $L_{\mathcal{K}}$ share the same parameters, i.e., the embedding vectors for words and their corresponding entities are the same. In the following subsections, we will introduce the text model to specify $L_{\mathcal{D}}$ and the knowledge model to specify $L_{\mathcal{K}}$.

\subsection{Text Model}
\label{text_loss}
Similar to \cite{RCNet,KnowTextJoint}, we leverage the Skip-Gram model \cite{NIPS2013W2V} as the text model. In Skip-Gram, the probability of observing the target word $w_O$ given its context word $w_I$ is modeled as
$P(w_O|w_I)=\frac{\exp(\mathbf{w'_O}\cdot\mathbf{w_I})}{\sum_{w\in\mathcal{V}}\exp(\mathbf{w'}\cdot\mathbf{w_I})}$,
where $\mathbf{w}\in\mathcal{R}^d$ and $\mathbf{w'}\in\mathcal{R}^d$ denote the input and output embedding vectors for word $w$ respectively, $\mathcal{V}$ is the dictionary, and $d$ is dimension of the embedding.

Given the training corpus $\mathcal{D}$ consisting of $|\mathcal{D}|$ token words $\{p_1,\cdots,p_k,\cdots, p_{|\mathcal{D}|}\}$, the loss $L_{\mathcal{D}}$ is specified by:
\begin{equation}
\label{eqn_text_loss}
L_{\mathcal{D}} = \sum_{k=1}^{|\mathcal{D}|}\sum_{j\in\{-M,\cdots,M\},j\neq 0}P(p_k|p_{k + j}),
\end{equation}
where $2M$ is the size of the sliding window. As it is expensive to directly minimize $L_{\mathcal{D}}$ due to the denominator of $P(w_O|w_I)$, we adopt the negative sampling strategy \cite{NIPS2013W2V} to boost the computation efficiency.

\subsection{Knowledge Model}
\label{Knowledge_loss}
The knowledge model in ProjectNet is based on an \emph{asymmetric low-rank projection} that projects the original entity embedding vectors into a new semantic space. The projection is designed to be \emph{asymmetric} in order to handle the heterogeneity between head and tail entities, and is designed to be \emph{low-rank} in order to deal with non \emph{one-to-one} relationships in the knowledge graphs.

\subsubsection{Asymmetric Projection}
As aforementioned, the head and tail entities in knowledge graphs are usually very different, from both semantic and statistical perspectives. Therefore, we argue that it is unreasonable to adopt the same projection to these two kinds of entities (as TransH \cite{TransH} does). Instead, it would be better to adopt different projection matrices, denoted as $L_r\in\mathcal{R}^{d\times d}$ and $R_r\in\mathcal{R}^{d\times d}$ respectively, to the head and tail entities. Hence, given a triple $(h,r,t)$, the original embedding vectors for $h$ and $t$ will be transformed to $\mathbf{h'}$ and $\mathbf{t'}$ as follows,
\begin{equation}
\label{eqn_assy_0}
\mathbf{h'}=L_r\mathbf{h},\ \mathbf{t'}=R_r\mathbf{t}.
\end{equation}
Based on the transformed embeddings, we define a  scoring function $f_d$ to reflect the confidence level that the triple $(h,r,t)$ is true:
\begin{equation}
\label{eqn_assy_1}
f_d(h,r,t)=||\mathbf{h'}+\mathbf{r}-\mathbf{t'}||_2^2 =||L_r\mathbf{h}+\mathbf{r}-R_r\mathbf{t}||_2^2.
\end{equation}
Then we adopt a margin based ranking loss to distinguish the golden relationship triples from randomly corrupted triples:
\begin{equation}
\label{eqn_know_loss}
\begin{aligned}
&L_{\mathcal{K}}=\sum_{(h,r,t)}\sum_{(h',r',t')\in N(h,r,t)}[\gamma - f_d(h',r',t') + f_d(h,r,t)]_+,
\end{aligned}
\end{equation}
where $[x]_+=\max(0,x)$, $\gamma>0$ is the margin value, $N(h,r,t)$ is the set of all the corrupted triples built for the triple $(h,r,t)$, and $L_r$ and $R_r$ will be specified in (\ref{eqn_low_rank}).

\subsubsection{Low-Rank Projection}
As mentioned in the introduction, many relationships in the knowledge graphs are non \emph{one-to-one}. In this case, in order to achieve reasonable results during the minimization of $ L_{\mathcal{K}}$ defined above, it is necessary to constrain the projection matrices $L_r$ and $R_r$ to be \emph{low-rank}, which is described in the following proposition.

\begin{proposition}
\label{prof:lowrank}
Once linear projections are imposed to head and tail entities, the necessary condition to overcome the non \emph{one-to-one} mapping problem is that the projection matrices $L_r$ and $R_r$ should not be full-ranked.

\begin{proof}
Consider the following least-square problem w.r.t. the optimization variable $\mathbf{h}$:
\begin{equation}
\min||L_r\mathbf{h}-\mathbf{c}||_2^2,
\end{equation}
where $L_r\mathbf{h} = \mathbf{h'}$ and we regard $\mathbf{c} = \mathbf{t'}-\mathbf{r}$ as a constant vector. It is easy to obtain that the optimal solution $\mathbf{h^*}$ satisfies the following linear system:
\begin{equation}
\label{eqn_satis}
L_r^TL_r\mathbf{h^*}=L_r^T\mathbf{c}.
\end{equation}

To avoid the non \emph{one-to-one} mapping problem, the above equation must have multiple solutions. Then it is necessary that $L_r^TL_r$ is a low-rank matrix. In addition, as $rank(L_r^TL_r)=rank(L_r)$, the linear projection matrix $L_r$ must not be full-rank either. The same conclusion holds for the projection matrix $R_r$ for the tail entity.
\qed
\end{proof}
\end{proposition}

Given the above proposition, we use the following tricks to ensure that $L_r$ and $R_r$ are low-rank matrices (whose ranks are $m_L$ and $m_R$ respectively, with $m_L<d$ and $m_R<d$):
\begin{equation}
\label{eqn_low_rank}
L_r=\sum_{i=1}^{m_L} \mu_{r}^{(i)}\mathbf{p_{r}^{(i)}}\mathbf{q_{r}^{(i)}}^T,\quad R_r=\sum_{i=1}^{m_R}\zeta_{r}^{(i)}\mathbf{o_{r}^{(i)}}\mathbf{s_{r}^{(i)}}^T,
\end{equation}
where $\mu_{r}^{(i)}$, $\zeta_{r}^{(i)}$ are scalars, and $\mathbf{p_{r}^{(i)}}$, $\mathbf{q_{r}^{(i)}}$, $\mathbf{o_{r}^{(i)}}$, $\mathbf{s_{r}^{(i)}}$ are all $d$-dimensional real vectors, the outer products of which constitute $(m_L + m_R)$ rank $1$ matrices $\mathbf{p_{r}^{(i)}q_{r}^{(i)}}^T$ and $\mathbf{o_{r}^{(i)}s_{r}^{(i)}}^T$. For simplicity, we set the rank of all the left matrices $L_r$ to be the same ($m_L$) and the rank of all the right matrices $R_r$ to be the same ($m_R$). Please note we can also specify different ranks for different relationships $r$. We leave the corresponding discussions to the future work.

\subsection{Discussions}
\label{relation}
In this section, we discuss the connections of our proposed ProjectNet algorithm with a few previous works and show that they are special cases of ProjectNet.

\textbf{RNet}. RNet refers to the knowledge models proposed in \cite{RCNet} and \cite{KnowTextJoint}. In fact, both models in the two works try to minimize the same scoring function: $f_d(h,r,t) = ||\mathbf{h} + \mathbf{r} -\mathbf{t}||_2^2$. Their only difference lies in how $f_d$ is minimized. In \cite{RCNet}, a large margin ranking loss is adopted for the minimization of $f_d(h,r,t)$, whereas in \cite{KnowTextJoint}, an approximate softmax loss is used.
It is clear that such a scoring function $f_d(h,r,t)$ cannot handle either the non \emph{one-to-one} relationships between entities or the heterogeneity between head and tail entities. To state it more formally, let us consider the relationship triples $(h_i,r,t),i\in{1,\cdots,N}$, where all head entities $h_i$ have the same relationship $r$ with tail entity $t$. In the ideal case, if all $f_d(h_i,r,t)$ are fully minimized, we will have $\mathbf{h_i}=\mathbf{t}-\mathbf{r}, \forall i\in{1,\cdots,N}$, which implies that $\mathbf{h_1}=\mathbf{h_2}=\cdots=\mathbf{h_N}$. It means that all the embedding vectors for the head entities $\{h_i\}_{i=1}^N$ are the same, which is clearly unreasonable. We may encounter similar issues for \emph{one-to-many} relationships $\{(h,r,t_j)\}_j$ and \emph{many-to-many} relationships $\{(h_i,r,t_j)\}_{i,j}$.

Note that RNet corresponds to $L_r=R_r=I_{d\times d}$ in (\ref{eqn_assy_0}) and since the identity matrix $I_{d\times d}$ can be written in the form of (\ref{eqn_low_rank}), RNet can be regarded as a special case of ProjectNet.

\textbf{TransH}.
TransH \cite{TransH} is proposed to overcome the non \emph{one-to-one} mapping problem. It first projects the entity embedding vectors $\mathbf{h}$ and $\mathbf{t}$ onto a hyperplane w.r.t the relationship $r$, and then the projected vectors $\mathbf{h_{\bot}}$ and $\mathbf{t_\bot}$ are used to define the scoring function $f_d$. Specifically,
\begin{equation}
\begin{aligned}
\label{eqn_transh}
\mathbf{h_\bot = h - w_r}^T\mathbf{hw_r},\ & \mathbf{t_\bot = t-w_r}^T\mathbf{tw_r},\\
f_d(h,r,t)=||&\mathbf{h_\bot}+\mathbf{r}-\mathbf{t_\bot}||_2^2,
\end{aligned}
\end{equation}
where $\mathbf{w_r}\in\mathcal{R}^d$ is the normal vector of the hyperplane with unit length (i.e., $\mathbf{w_r}\cdot\mathbf{r}=0$ and $||\mathbf{w_r}||_2=1$).

Our proposed ProjectNet model differs from TransH  in two ways: (i) we adopt different projections to head and tail entities; (ii) we adopt general projection matrices rather than a hyperplane based projection. Actually, TransH (\ref{eqn_transh}) can be regarded as a special case of ProjectNet (\ref{eqn_know_loss}), as shown below. Starting from (\ref{eqn_transh}), we have
\begin{equation}
\mathbf{h_{\bot}}=\mathbf{h}-\mathbf{w_r}^T\mathbf{hw_r}=\mathbf{h}-\mathbf{w_r}\mathbf{w_r}^T\mathbf{h}=(I-\mathbf{w_rw_r}^T)\mathbf{h}.
\end{equation}
Hence, by substituting $L_r=(I-\mathbf{w_rw_r}^T)$ in (\ref{eqn_assy_0})(\ref{eqn_assy_1}), we get TransH. We still need to check whether $L_r=(I-\mathbf{w_rw_r}^T)$ can be written in the form of (\ref{eqn_low_rank}), i.e., the weighted sum of $m_L$ rank-$1$ matrices, where $m_L<d$. We answer this question in the following two steps:
(i)	As $L_r=(I-\mathbf{w_rw_r}^T)$ is an idempotent matrix (i.e. $L_rL_r=L_r$) and $\mathbf{w_r}$ is a unit length vector, it holds that $rank(L_r)=trace(L_r)=d-1$ \cite{Matrix}. Therefore, $L_r$ has $d-1$ non-zero eigenvalues. Furthermore, by observing that the eigenvalues of $\mathbf{w_rw_r}^T$ are $0$ and $1$, we can conclude that $L_r=(I-\mathbf{w_rw_r}^T)$ has $1$ as one of its eigenvalues, corresponding to $d-1$ linearly independent eigenvectors, and $0$ as its another eigenvalue, corresponding to one eigenvector.
(ii) Further considering that $L_r$ is a real symmetric matrix, we can decompose $L_r$ as $L_r=U_r\Sigma_r U_r^T$, where $U_r=(\mathbf{u_r^{(1)}},\mathbf{u_r^{(2)}},\cdots,\mathbf{u_r^{(d)}})\in\mathcal{R}^{d\times d}$ and $\Sigma_r=diag(1,1,\cdots,1,0)\in\mathcal{R}^{d\times d}$. The first $d-1$ columns $\{\mathbf{u_r^{(i)}}\}_{i=1}^{d-1}$ of $U_r$ are all the unit-length eigenvectors of $L_r$ corresponding to eigenvalue $1$ and $\Sigma_r$ stores all the eigenvalues of $L_r$. Thus we can write $L_r=\sum_{i=1}^{d-1}\mathbf{u_r^{(i)}u_r^{(i)}}^T$.

The same procedure holds for the relation between $R_r\mathbf{t}$ and $\mathbf{t_\bot}$. Then according to the above discussions, we can obtain the following proposition.
\begin{proposition}
In the knowledge model of ProjectNet (\ref{eqn_low_rank})(\ref{eqn_assy_1}), letting $m_L =m_R =d - 1$, $\mu_{r}^{(i)}=\zeta_{r}^{(i)}=1$ and $\mathbf{p_r^{(i)}}=\mathbf{q_{r}^{(i)}}=\mathbf{o_{r}^{(i)}}=\mathbf{s_{r}^{(i)}}=\mathbf{u_r^{(i)}}$, where $\mathbf{u_r^{(i)}}$ is the $i^{th}$
eigenvector of the matrix $I-\mathbf{w_rw_r}^T$ with unit length, $i=1,2,\cdots,d-1$, we can obtain the TransH model (\ref{eqn_transh}).
\end{proposition}

\textbf{SE}.
SE \cite{BordesAAAI} adopts the following scoring function:
\begin{equation}
f_d(h,r,t)=||L_r\mathbf{h}-R_r\mathbf{t}||_1.
\end{equation}
SE looks very similar to our proposed knowledge model. However, there is a key difference: SE does not add the low-rank constraints to the matrices $L_r$ and $R_r$. In other words, they fix the rank of these two matrices to be full, while in our model the rank of the matrices is a variable. Therefore our model is more general than SE and can handle the non \emph{one-to-one} relationship when the rank is low while SE cannot since its rank is always full. In this sense, we could also regard SE as a special case of our proposed ProjectNet model.

\textbf{TransR}.
\cite{TransR} TransR treats relationships and entities as different objects and thus separates their embeddings into different spaces,
\begin{equation}
f_d(h,r,t)=||M_r\mathbf{h}+\mathbf{r}-M_r\mathbf{t}||_2^2.
\end{equation}
Different with our formulation (\ref{eqn_low_rank})(\ref{eqn_know_loss}), they did not add the low-rank constraint to matrix $M_r$ (or we say that it sets the matrix $M_r$ to be full-rank). In addition, TransR adopts the same transformation matrices to head and tail entities, by assuming that they are located in the same space. Therefore, the knowledge model in our ProjectNet algorithm is more general than TransR, and can include it as our special case.

\section{Experiments}
\label{exp}
In this section, we conduct a set of experiments to verify the effectiveness of the ProjectNet model.

\subsection{Experiments Setup}

\subsubsection{Training Data}
\label{train_data}
For the free text corpus, we used a public snapshot of English Wikipedia named \emph{enwik9}.\footnote{http://mattmahoney.net/dc/enwik9.zip} The corpus contains about $120$ million word tokens. We removed digital words and words with frequency less than $5$. Then we leveraged a knowledge graph \emph{FB13} \cite{Socher2013NTN} to impose relationships onto those entities covered by \emph{enwik9}. Since \emph{FB13} contains many entities whose names have multiple words, in \emph{enwik9} we merged these words into phrases and regarded both single words and phrases as embedding units in the dictionary. Finally the dictionary size is about $230k$.

\subsubsection{Baseline Methods}

We consider the following algorithms as our baselines (we used the codes released by the authors of these works for implementation):
\begin{enumerate}
  \item \textbf{Skip-Gram (SG)}: the original Skip-Gram model in \emph{word2vec}, corresponding to $\alpha=0$ in (\ref{eqn_total_loss}).
  \item \textbf{RNet}: the joint embedding model in \cite{RCNet} and \cite{KnowTextJoint}, which adopts the objective $\min||\mathbf{h}+\mathbf{r}-\mathbf{t}||_2^2$ in the knowledge model.\footnote{As aforementioned, the models in the two papers differ only in the loss function (i.e., ranking loss vs. approximate softmax loss). Hence, we unify these two models using the name RNet and report the better performance of the two loss functions.}
  \item \textbf{Skip-Gram+TransH (SG+TransH)}: the combination of Skip-Gram (for the text model) and TransH (for the knowledge model). According to the discussions in Section \ref{relation}, this baseline is a special case of ProjectNet.
\end{enumerate}

\subsubsection{Parameter Setting}
\label{parameter}
In our experiments, we set the embedding size to $d=100$. Stochastic Gradient Descent(SGD) is used to train all the models. We set the initial learning rate to be $0.025$ and linearly dropped it during the training process. For the knowledge model in ProjectNet, we initialized the projection matrices $L_r$ and $R_r$ to be diagonal matrices with randomly assigned $0, 1$ elements (with $m_L$ and $m_R$ non-zero elements respectively). For $m_L$ and $m_R$, we varied their values according to the set $\{10,20,\cdots,80,90,95,100\}$. For all the joint embedding models, we varied the trade-off parameter $\alpha$ in (\ref{eqn_total_loss}) according to the set $\{0.01,0.05,0.1,0.2,0.5\}$. The margin value is set to $\gamma=1$.

We used two tasks to evaluation our algorithm and the baseline models, one is the analogical reasoning task and the other is the word similarity task. The corresponding experiments results are shown in the following two subsections.

\subsection{Analogical Reasoning Task}
\label{task_ana}

\begin{table*}[ht]
\setlength{\abovecaptionskip}{+0.1cm}
\setlength{\belowcaptionskip}{-0.4cm}
\footnotesize
\centering
\begin{tabular}{|c|c|c|c|c|c|}
\hline
Relationship & \#Question &\textbf{SG}&\textbf{RNet}&\textbf{SG+TransH}&\textbf{ProjectNet} \\ \hline
\emph{cause\_of\_death} & 4290 &3.29\% &4.31\%&7.55\% &10.84\% \\ \hline
\emph{nationality}& 870 &14.60\%&14.14\% &14.82\% &17.47\% \\ \hline
\emph{gender} & 650 & 67.08\%&59.38\%&75.54\% &84.62\%\\ \hline
\emph{profession} & 6320 & 3.73\%&5.78\%& 8.66\%&13.42\%\\ \hline
\emph{institution} & 4556 & 1.54\%&3.03\%&4.92\% &6.72\%\\ \hline
\emph{ethnicity} & 342 & 16.96\%&15.50\%&15.79\% &18.71\%\\ \hline
\emph{religion} & 3192 & 13.00\%&12.47\%&15.88\% &22.06\%\\ \hline
\textbf{Total} & \textbf{20220} & \textbf{7.33\%}&\textbf{8.15\%}&\textbf{11.24}\% &\textbf{15.28\%}\\ \hline
\end{tabular}
\caption{Accuracy of different models on analogical reasoning task.}
\label{tbl_ana_result}
\end{table*}

The analogical reasoning task is a word relationship inference task proposed in \cite{NIPS2013W2V}. It consists of several quadruple word questions \emph{a:b,c:d}, in which the relationship between word \emph{a} and \emph{b} is the same as that between \emph{c} and \emph{d}. For instance, $(a:b,c:d) = (Berlin : Germany , Paris : France)$ and the relationship $r$ is \emph{capital-countries}. The task aims to infer word \emph{d} given words \emph{a}, \emph{b}, and \emph{c} using their word embedding vectors. To be more concrete, the inferred word $\hat{d}$ is given by $\hat{d}=\arg\max_{w\in\mathcal{V}} cosine(\mathbf{b}-\mathbf{a}+\mathbf{c},\mathbf{w})$.  Once $\hat{d}=d$, the result on this quadruple word question is right; otherwise, it is wrong.

To construct the test set for the analogical reasoning task, we randomly sampled $1\%$ triples from \emph{FB13}, and filtered them according to the dictionary of \emph{enwik9}.\footnote{We did not use the analogical reasoning dataset given in \cite{NIPS2013W2V} because this dataset is too special in the sense that almost all the relationships in it are \emph{one-to-one} mappings.}
This test set consists of about $20k$ questions belonging to $7$ non \emph{one-to-one} relationships. The detailed statistics for this test dataset can be found in Table \ref{tbl_ana_result}.

Then, we went through the remaining triples in \emph{FB13} and removed all those triples containing overlapped entities with the test data. In this way, we obtained a training set with roughly $76k$ triples, which has no overlap with the test set in either relationship triples or entities. The goal of doing so is to examine whether the free text corpus can act as a bridge between known and unknown entities, so as to verify the necessity of jointly embedding text and knowledge into the distributed representation space.

For ProjectNet, as we imposed $L_r\mathbf{a}-R_r\mathbf{b}=-\mathbf{r}=L_r\mathbf{c}-R_r\mathbf{d}$ instead of $\mathbf{a}-\mathbf{b}=\mathbf{c}-\mathbf{d}$, we took a two-step approach instead of directly using the original word vectors to perform the analogical reasoning task: (i) we chose an optimal relationship $r^*$ that best describes the relationship between \emph{a} and \emph{b}, i.e., $r^*=\arg\min_r||L_r\mathbf{a}+\mathbf{r}-R_r\mathbf{b}||_2^2$; (ii) under $r^*$, we chose the answer word $\hat{d}$ according to $\hat{d}=\arg\min_{w\in\mathcal{V}}||L_{r^*}\mathbf{c}+\mathbf{r^*}-R_{r^*}\mathbf{w}||_2^2$. The same evaluation method was also applied to SG+TransH as well.

The experimental results are listed in Table \ref{tbl_ana_result}, from which we have the following observations:
\begin{itemize}
\item All the knowledge based models (RNet, SG+TransH, and ProjectNet) outperform the original SG model, indicating that the quality of word embedding can be improved by leveraging knowledge graphs.
\item The two models that take non \emph{one-to-one} mappings into consideration (i.e., SG+TransH and ProjecNet) are superior to RNet, showing the necessity of modeling the non \emph{one-to-one} mappings into the loss functions.
\item Among all the models, ProjectNet achieves the best performance in all the seven subtasks. For the overall accuracy, it achieves over $30\%$ relative gain than SG+TransH. This well demonstrates the advantages of our proposed model.
\end{itemize}

\subsubsection{Sensitivity to different ranks}
\begin{figure}[ht]
\setlength{\abovecaptionskip}{+0.1cm}
\setlength{\belowcaptionskip}{-0.35cm}
	\centering
	\scalebox{0.46}[0.44]{\includegraphics{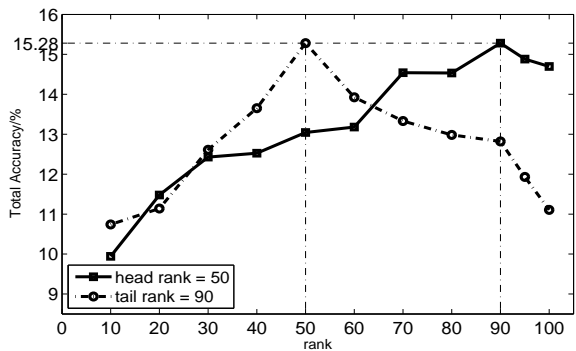}}
    \footnotesize
	\caption{Accuracy w.r.t. different head and tail ranks. The dashdot line records the accuracy varying with different head ranks when tail rank is fixed to 90. The solid line records the accuracy varying with different tail ranks when head rank is fixed to 50.}\label{fig:lda_plage}
    \label{fig_rank_survey}
\end{figure}

The best performance of ProjectNet was obtained with $\alpha=0.2$, $m_L=50$, and $m_R=90$. To show the influence of the ranks of the projection matrices, in Figure \ref{fig_rank_survey}, we plotted two curves that reflect the performance of ProjectNet w.r.t. different rank values $m_L$ and $m_R$: one curve corresponds to changing $m_R$ while fixing $m_L=50$, and the other corresponds to changing $m_L$ while fixing $m_R=90$. From the figure, we have the following observations. (i) The performance becomes bad when the rank is too low. This is because in this case the model expressiveness becomes poor due to small number of free parameters in the projection matrices. (ii) For the projection matrix for head entities,  medium values of $m_L$ correspond to the better performances (the dashdot line), while for the projection matrix for tail entities, higher values of $m_R$ lead to better performances (the solid line). This result is consistent with the statistical information in Table \ref{tbl_statis_result}: the degree of non \emph{one-to-one} mappings for head entities is much higher than that for tail entities, suggesting a lower rank of projection matrix for head entities.

\subsection{Word Similarity Task}
Word similarity is a task to investigate whether the similarity computed from word embedding vectors is consistent with human-labeled word similarity. We used three word-similarity tasks in our experiments, namely Word Similarity $353$ (WS353) \cite{WS353}, SCWS \cite{EricHuang2012} and Rare Word (RW) \cite{Luong13_morpheme}. There are $353$, $2003$, and $2034$ word pairs in these datasets respectively. From the word embedding vectors, we obtain the similarity scores (e.g., cosine similarity) for each word pair, based on which a ranked list is derived on the word pairs. Then the generated ranked list is compared to the ranked list produced by the ground-truth similarity scores assigned by human labelers. To evaluate the consistency between two ranking lists, we used Spearman's Rank Correlation (denoted as $\rho\in[-1,1]$). Higher $\rho$ corresponds to better word embedding vectors.

Table \ref{tbl_ws_result} summarizes the results. For ProjectNet and SG+TransH, the word embedding vectors were directly used to compute the similarity scores, which is different from the analogical reasoning task. This is because there is no explicit relationship available in the evaluation process. The best performances of ProjectNet on the three datasets were obtained with the parameters setting to $(m_L=50, m_R=95, \alpha=0.05)$, $(m_L=40, m_R=90, \alpha=0.01)$, and $(m_L=60, m_R=95, \alpha=0.05)$ respectively. Table \ref{tbl_ws_result} reveals that ProjetNet achieves the best performance on all the datasets, which further indicates that ProjetNet produce higher quality word embedding vectors than the baseline methods.

\begin{table}[htbp]
\setlength{\abovecaptionskip}{+0.1cm}
\setlength{\belowcaptionskip}{-0.4cm}
\footnotesize
\centering
\begin{tabular}{|c|c|c|c|c|}
\hline
Task/Model &\textbf{SG}&\textbf{RNet}&\textbf{SG+TransH}&\textbf{ProjectNet} \\ \hline
\emph{WS353}  &0.647&0.661&0.666 &\textbf{0.684} \\ \hline
\emph{SCWS} &0.610&0.614&0.618 &\textbf{0.630}\\ \hline
\emph{RW} &0.179 &0.184 &0.187 &\textbf{0.198} \\ \hline
\end{tabular}
\caption{Spearman's Rank Correlation($\rho$) on three Word Similarity Datasets: WS353, SCWS, and RW. Each $\rho$ is reported as the average value of five repeated runs.}
\label{tbl_ws_result}
\end{table}

\section{Conclusions and Future Work}
\label{Conclusion}
In this paper we proposed a novel word embedding algorithm called ProjetNet, which leverages knowledge graphs to improve the quality of word embedding. In ProjetNet, we adopt different asymmetric low rank projections to head and tail entities in an entity-relationship triple, thus successfully maintain both non \emph{one-to-one} mapping and heterogenous head/tail entities properties of knowledge graph. Experimental results demonstrate that ProjetNet significantly outperforms previous embedding models.

For the future work, we plan to apply the proposed approach to fulfill knowledge mining tasks, such as triplet classification and link prediction \cite{TransH}. In addition, we plan to use the word embedding vectors generated by ProjectNet in some real-world applications such as document classification and web search ranking.

\bibliographystyle{plain}
\bibliography{proj_net}

\end{document}